%% file: main.tex
\documentclass[letterpaper, 10 pt, conference]{ieeeconf}  % Comment this line out if you need a4paper

\IEEEoverridecommandlockouts                              % This command is only needed if 
\usepackage{xcolor} 
\usepackage{graphicx}
\usepackage{booktabs}
\usepackage{multirow}
\usepackage{array}
\usepackage{cuted}
\usepackage{stfloats}
\usepackage{amsmath} % in your preamble, if not already
\usepackage{cuted}
\usepackage{afterpage}
\usepackage{capt-of} % For \captionof
\usepackage{etoolbox} % For \patchcmd or \apptocmd

\makeatletter
\patchcmd{\@maketitle}{\null}{{\myfigure{}\par}}{}{}
\makeatother
\usepackage{pifont}
\usepackage{amssymb}
\usepackage{hyperref} 
\title{\LARGE \bf
Humanoid Agent via Embodied Chain-of-Action Reasoning with Multimodal Foundation Models for Zero-Shot Loco-Manipulation \vspace{0pt}}

\author{Congcong Wen$^{*}$, Geeta Chandra Raju Bethala$^{*}$, Yu Hao, Niraj Pudasaini, Hao Huang, \\ Shuaihang Yuan,   Baoru Huang, Anh Nguyen, Mengyu Wang, Anthony Tzes, Yi Fang% 
\thanks{Congcong Wen, Geeta Chandra Raju Bethala, Yu Hao, Niraj Pudasaini, Hao Huang, Shuaihang Yuan, and Yi Fang are with Embodied AI and Robotics (AIR) Lab, New York University, New York, USA and NYUAD Center for Artificial Intelligence and Robotics, New York University Abu Dhabi, Abu Dhabi, UAE. Congcong Wen is also with the Harvard AI and Robotics Lab, Harvard University, Boston, USA.
{\tt\small \{cw3437, gb2643, yh3252, hh1811, sy2366, np2289, yf23\}@nyu.edu}}
\thanks{Baoru Huan is with the  the Department of Computer Science, University College
London, London, UK. {\tt\small baoru.huang@ucl.ac.uk }}
\thanks{Anh Nguyen is with the Department of Computer Science, University of Liverpool, UK. {\tt\small anh.nguyen@liverpool.ac.uk }}
\thanks{Mengyu Wang is with the Harvard AI and Robotics Lab, Harvard University, Boston, USA. {\tt\small mengyu\_wang@meei.harvard.edu}}%
\thanks{Anthony Tzes is with the NYUAD Center for Artificial Intelligence and Robotics, New York University Abu Dhabi, Abu Dhabi, UAE. {\tt\small anthony.tzes@nyu.edu}}%
\thanks{$^{*}$These authors contributed equally; order determined by a coin toss.}%  
}

\begin{document}

\maketitle

% \begin{strip}
% \vspace{-55pt}
% \centering
% \includegraphics[width=\textwidth]{imgs/tasks.jpg}
% \vspace{-20pt}
% % \captionof{figure}{Real-world qualitative results on three humanoid loco-manipulation tasks.}
% \captionof{figure}{Real-world qualitative results on three humanoid loco-manipulation tasks. In the \textbf{top row (Loco-Manipulation)}, the robot holds a box, locates the target, walks to it, and places the box. The \textbf{middle row (Object Manipulation)} the robot identifies a bottle, grasps it, lifts it, and places it. In the \textbf{ bottom row (Rearrangement and Manipulation)} robot locates multiple objects, rearranges them, and places the selected object in the designated position.}
% \label{fig:humanoid_tasks}
% \vspace{-10pt}

% \end{strip}

\input{sec/0_abstract}

\input{sec/1_intro}

\input{sec/2_relatedwork}

\input{sec/3_method}

\input{sec/4_experiments}

\input{sec/5_conclusion}

%\section*{APPENDIX}

% \section*{ACKNOWLEDGMENT}

% The preferred spelling 

%%%%%%%%%%%%%%%%%%%%%%%%%%%%%%%%%%%%%%%%%%%%%%%%%%%%%%%%%%%%%%%%%%%%%%%%%%%%%%%%
% \clearpage

% {
%     \small
%     % \bibliographystyle{ieeenat_fullname}
%     % \bibliographystyle{ieeetr}
%     \bibliography{main}
    
% }

\input{main.bbl}
\end{document}

%% file: sec/0_abstract.tex
\begin{abstract}

Humanoid loco-manipulation, which integrates whole-body locomotion with dexterous manipulation, remains a fundamental challenge in robotics. Beyond whole-body coordination and balance, a central difficulty lies in understanding human instructions and translating them into coherent sequences of embodied actions. Recent advances in foundation models provide transferable multimodal representations and reasoning capabilities, yet existing efforts remain largely restricted to either locomotion or manipulation in isolation, with limited applicability to humanoid settings. In this paper, we propose Humanoid-COA, the first humanoid agent framework that integrates foundation model reasoning with an Embodied Chain-of-Action (CoA) mechanism for zero-shot loco-manipulation. Within the perception–reasoning–action paradigm, our key contribution lies in the reasoning stage, where the proposed CoA mechanism decomposes high-level human instructions into structured sequences of locomotion and manipulation primitives through affordance analysis, spatial inference, and whole-body action reasoning. Extensive experiments on two humanoid robots, Unitree H1-2 and G1, in both an open test area and an apartment environment, demonstrate that our framework substantially outperforms prior baselines across manipulation, locomotion, and loco-manipulation tasks, achieving robust generalization to long-horizon and unstructured scenarios. Project page: \href{https://humanoid-coa.github.io/}{https://humanoid-coa.github.io/}

%including ablation studies that assess the effectiveness of the chain of robotic action reasoning strategies in comprehending complex human instructions.. 
%Our method systematically decomposes complex tasks into transparent, intermediate steps, significantly enhancing the model's interpretability.
%This approach not only boosts the model's ability to handle complex and multi-step instructions but also enhances interpretability.

\end{abstract}

%% file: sec/1_intro.tex
\section{INTRODUCTION}
\label{sec:intro}

Humanoid loco-manipulation, which combines whole-body locomotion with dexterous manipulation, has long been recognized as a fundamental challenge in robotics. The difficulty arises not only from coordinating high-dimensional degrees of freedom and maintaining dynamic balance, but also from the cognitive demand of grounding human instructions into coherent sequences of embodied actions. While the former challenge has been substantially advanced by recent progress~\cite{h2-compact} in maintaining stability under external loads during loco-manipulation, the latter remains more fundamental: a central difficulty lies in bridging high-level human instructions with the low-level motor control required to realize complex whole-body actions in long-horizon and unstructured environments.

To address this cognitive challenge, early work focused on semantic mapping, where natural language instructions were translated into symbolic task representations or semantic maps that could be executed by motion planners~\cite{tellex2011understanding, matuszek2013learning}. Subsequent studies shifted toward language-conditioned policies that directly ground instructions into robot actions through imitation or reinforcement learning~\cite{andreas2017modular, sharma2021skill}. While both directions demonstrated effectiveness in constrained domains, they relied heavily on manual annotation and predefined task structures, limiting their scalability and adaptability.

Foundation models have recently emerged as a powerful paradigm for robotics, offering transferable multimodal representations and reasoning capabilities across diverse tasks and embodiments~\cite{li2023blip,2023gpt4}. Early efforts such as SayCan~\cite{ahn2022can} and PaLM-E~\cite{driess2023palm} demonstrated how large language and vision–language models can ground natural language instructions in robotic affordances, combining high-level reasoning with low-level motor control. Building on this foundation, subsequent works applied large-scale vision–language–action models to concrete domains such as zero-shot object-goal navigation~\cite{wen2025zero,huang2024gamap}, social navigation~\cite{wen2024socially}, and object grasping~\cite{vuong2024grasp}, showing promising generalization beyond task-specific training. Nevertheless, these efforts remain largely confined to either locomotion or manipulation in isolation, with most evaluations conducted in simulation or on non-humanoid platforms, thereby limiting their applicability to the high-dimensional and tightly coupled challenges of humanoid loco-manipulation. More recently, Wang et al.~\cite{wang2024autonomous} proposed an LLM-based behavior planning framework that leverages a grounded language model and a predefined behavior library to generate task graphs with integrated failure recovery. However, such approaches still underutilize the reasoning capabilities of LLMs, often relying on relatively direct mappings between instructions and actions.

In this paper, we address these limitations by introducing a humanoid agent framework, Humanoid-COA, which incorporates an Embodied Chain-of-Action (CoA) Reasoning mechanism for zero-shot loco-manipulation. Our framework is built upon the classical perception–reasoning–action paradigm, with its core innovation lying in the reasoning stage. Specifically, CoA Reasoning incrementally decomposes high-level instructions into structured sequences of locomotion and manipulation primitives. Unlike prior methods that rely on direct mappings or fixed task templates, CoA Reasoning integrates three complementary processes: object affordance analysis to identify actionable object properties, region spatial reasoning to infer occluded or unseen entities, and whole-body action reasoning to ensure kinematic and dynamic feasibility. This mechanism enables the agent to bridge human instructions with physically realizable trajectories in long-horizon and unstructured environments. Our main contributions are as follows:
\begin{itemize}
    \item To the best of our knowledge, we present the first \textit{humanoid agent framework} that integrates foundation model reasoning for zero-shot loco-manipulation under natural language instructions.
    \item We propose an \textit{Embodied Chain-of-Action Reasoning} mechanism that enables the humanoid agent to decompose high-level human intent into executable whole-body behaviors for long-horizon tasks in unstructured environments.
    \item We demonstrate through extensive experiments on two humanoid robots, including Unitree H1 and Unitree G1, that our framework achieves robust zero-shot generalization across diverse loco-manipulation tasks, substantially outperforming prior approaches.
\end{itemize}

\begin{figure*}
  \centering
  \includegraphics[width=0.95\textwidth]{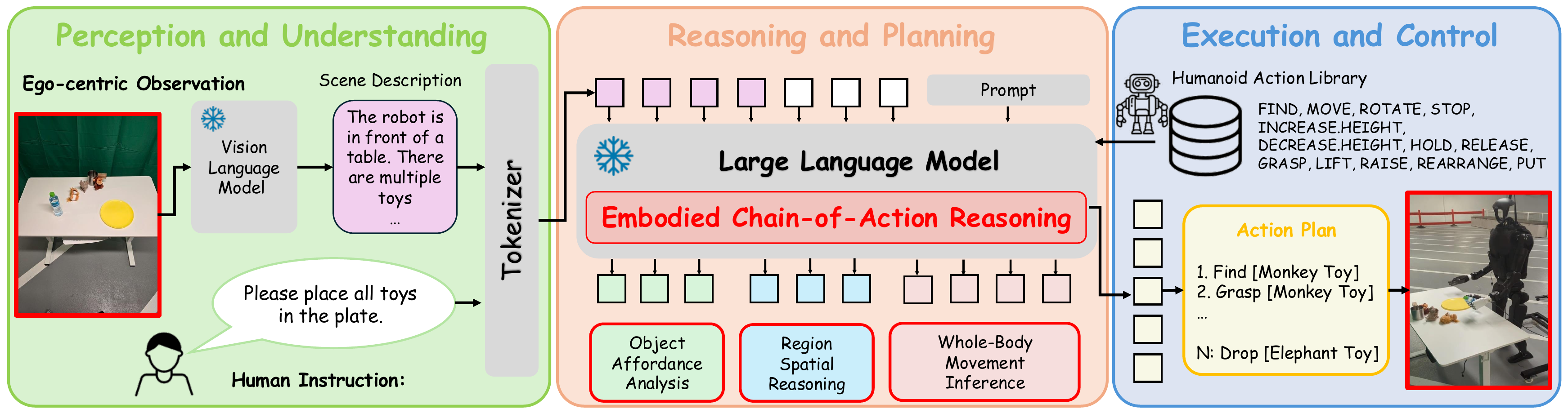}
 \caption{The proposed Humanoid Agent Framework for loco-manipulation, consisting of three stages: (i) Perception and Understanding, where ego-centric observations are converted into scene descriptions and, together with human instructions, tokenized for reasoning; (ii) Reasoning and Planning, where a large language model with \textit{Embodied Chain-of-Action Reasoning} generates symbolic action plans via affordance, spatial, and whole-body inference; and (iii) Execution and Control, where plans are grounded into primitive commands and translated into low-level motor control for humanoid execution.} \vspace{-20pt}
  \label{fig:overview}
\end{figure*}

%% file: sec/2_relatedwork.tex
\section{RELATED WORKS}
\label{sec:related_work}

\subsection{Foundation Models in Robotics}

Foundation models have recently emerged as a powerful paradigm for robotics, offering transferable representations and reasoning capabilities across diverse tasks and embodiments~\cite{2023gpt4,lin2025rs}. Early efforts such as SayCan~\cite{ahn2022can}, Inner Monologue~\cite{huang2022inner}, and PaLM-E~\cite{driess2023palm} demonstrated how large language or vision–language models can ground natural language instructions into robotic skills, combining high-level reasoning with low-level control. Building on these advances, subsequent works have applied foundation models to concrete tasks such as zero-shot object-goal navigation~\cite{wen2025zero,huang2024gamap}, social navigation~\cite{wen2024socially}, and object grasping~\cite{vuong2024grasp}, showing promising generalization in simulated benchmarks or environments. Nevertheless, these efforts remain largely confined to isolated locomotion or manipulation tasks, and their experiments are typically conducted in simulation or on non-humanoid platforms, limiting their applicability to the high-dimensional challenges of humanoid loco-manipulation. To address this gap, we propose a {humanoid agent framework with CoA Reasoning mechanism}, extending foundation model reasoning to the more complex setting of humanoid loco-manipulation.

\subsection{Loco-manipulation in Humanoids}
Humanoid loco-manipulation is inherently challenging as it requires coordinating high-dimensional degrees of freedom while maintaining balance and achieving task-oriented manipulation objectives~\cite{sferrazza2024humanoidbench}. Traditional approaches have primarily relied on model-based whole-body control frameworks~\cite{khatib_wb}, which decouple locomotion and manipulation through hierarchical optimization under physical constraints. While effective in structured settings, these methods depend on precise modeling and struggle with adaptability in unstructured environments. Recent advances in learning-based locomotion~\cite{kumar2021rma,smith2022legged} and combined locomotion–manipulation, mostly demonstrated on quadruped platforms~\cite{sun2022fully,9561835}, show encouraging results but remain limited in addressing bi-pedal and bi-manual humanoids. Efforts on humanoid platforms~\cite{wang2023physhoi,xie2023hierarchicalplanningcontrolbox} have achieved promising demonstrations, yet these are often constrained to specific tasks rather than generalizable frameworks. More recently, Wang et al.~\cite{wang2024autonomous} introduced an LLM-based behavior planning method that leverages a grounded language model and a predefined behavior library to generate task graphs with integrated failure recovery. However, such approaches only partially exploit the reasoning capacity of LLMs, relying mainly on direct mappings from instructions to actions. In contrast, we propose a humanoid agent framework with CoA Reasoning, which explicitly harnesses LLM reasoning to enable robust and interpretable action planning for zero-shot loco-manipulation in unstructured environments.

% In parallel, HYPERmotion~\cite{wang2024hypermotion} combines reinforcement-learned motion primitives, whole-body optimization, and task graphs generated by language models to enable hybrid loco-manipulation in humanoids. 

%% file: sec/3_method.tex
\section{Method}
% In this paper, we propose an integrated approach using embodied chain of action reasoning for humanoid loco-manipulation, based on a multi-modal foundation model as shown in Figure~\ref{fig2}. %In Section~\ref{method1}, we define the problem of humanoid loco-manipulation where, based on observations and a predefined humanoid action library, the robot infers human instructions to output an optimal list of actions. Section~\ref{method2} elaborates on our proposed chain of thought reasoning process, which encompasses the analysis of target object affordance, the management of spatial reasoning for unseen and occluded objects, and the inference of necessary body movements for task execution. In Section~\ref{method3}, we details the Humanoid Action Library, specifying the range of actions available to the robot.
\subsection{Problem Definition}
\label{problem}

We formalize the problem of {humanoid loco-manipulation}, which integrates locomotion (whole-body mobility) and manipulation (object interaction), as the task of generating executable action sequences in complex and unstructured environments. Formally, it is defined as follows: given a natural language instruction $I$, ego-centric observations $O$ of the environment, and a predefined humanoid action library $\mathcal{L} = \{\pi_1, \pi_2, \dots, \pi_n\}$ consisting of primitive skills (e.g., moving, grasping, raising and lifting), the objective is to produce an action sequence
\begin{equation}
    A = \{a_1, a_2, \dots, a_T\}, \quad a_t \in \mathcal{L},
\end{equation}
that fulfills the task specified by $I$ under the physical and dynamical constraints of the robot. This can be formalized as learning a mapping function
\begin{equation}
    f : (I, O, \mathcal{L}) \mapsto A.
\end{equation}

The fundamental challenge lies in bridging the gap between \textit{high-level abstract instructions} and \textit{low-level embodied execution}, particularly under conditions of partial observability and dynamically changing environments. To address this gap, we introduce a {humanoid agent framework} that leverages multi-modal foundation models and embodied chain-of-action reasoning to synthesize executable whole-body action plans in a zero-shot manner.

\begin{figure}[t]
  \centering
  \includegraphics[width=0.5\textwidth]{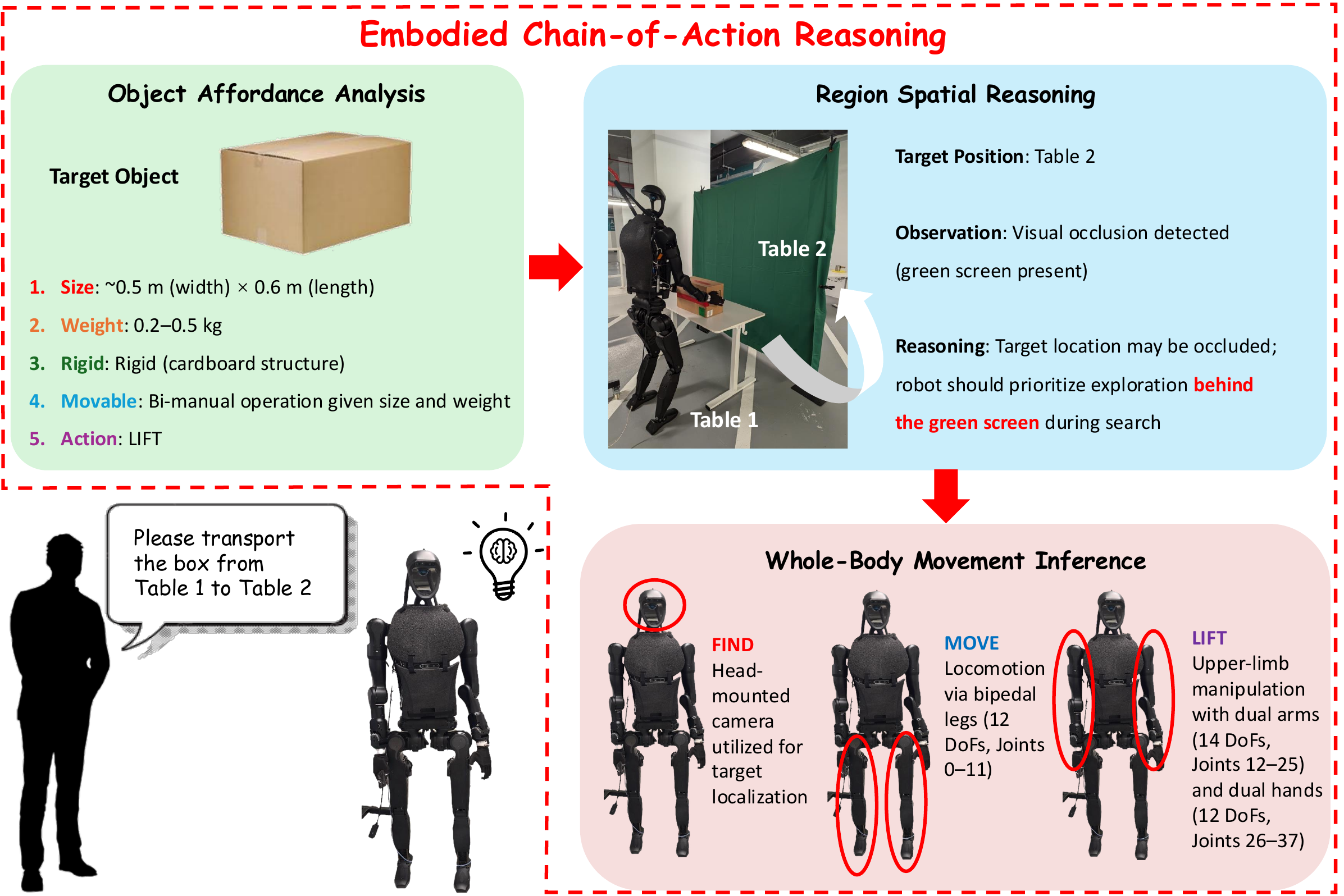}
 \caption{Example of the proposed Embodied Chain-of-Action Reasoning. Given a natural language instruction, the framework sequentially performs \textit{Object Affordance Analysis} to extract target properties and feasible actions, \textit{Region Spatial Reasoning} to handle occlusion and prioritize search areas, and \textit{Whole-Body Movement Inference} to map symbolic primitives onto the humanoid’s sensorimotor system. } \vspace{-20pt}
  \label{fig:COA}
\end{figure}

\subsection{Humanoid Agent Framework}
\label{method2}

We present a {humanoid agent framework}, Humanoid-COA, for zero-shot loco-manipulation that follows the classical \emph{perception–reasoning–action} paradigm (Fig.~\ref{fig:overview}). In the \emph{perception and understanding} stage, the agent integrates multimodal sensory inputs to capture geometric structures, semantic attributes, and affordance cues, while grounding natural language instructions into structured task representations aligned with the perceived environment. The core of our framework lies in the \emph{reasoning and planning} stage, where we introduce {Embodied Chain-of-Action (CoA) Reasoning}. This mechanism decomposes high-level task goals into a structured sequence of loco-manipulation primitives through three complementary processes: (i) \emph{object affordance analysis}, which assesses physical properties such as size, weight, rigidity, and movability to determine feasible object-level actions; (ii) \emph{region spatial reasoning}, which infers the existence and plausible locations of occluded or unseen targets and guides exploration accordingly; and (iii) \emph{whole-body movement inference}, which evaluates the feasibility of coordinated locomotion and manipulation under kinematic and dynamic constraints, instantiating primitives such as \texttt{FIND}, \texttt{MOVE}, and \texttt{LIFT}. Finally, in the \emph{action execution} stage, the whole-body controller instantiates the planned action chain into motor commands, enabling robust interaction with unstructured environments.

\subsubsection{Perception and Understanding}

The perception and understanding stage prepares both the environmental context and the task intent in textual form, serving as the input to the reasoning module. Given an ego-centric RGB observation $O \in \mathbb{R}^{H \times W \times 3}$, we employ a pre-trained {vision--language foundation model} (VLM) $f_{\text{vlm}}$, trained on large-scale image--text pairs, to translate the raw visual input into a natural-language {scene description}. Formally, \( S = f_{\text{vlm}}(O),\) where $S$ captures objects, attributes, and spatial relations in free-form text, enabling open-vocabulary and context-aware description beyond the closed-set labels of conventional object detectors. In parallel, the task objective is specified through a natural-language instruction $I$ provided by the user, which conveys the high-level goal to be accomplished by the agent. Then both the scene description $S$ and the instruction $I$ are tokenized into discrete sequences:
\begin{equation}
    {T}_S = f_{\text{Tokenizer}}(S), \qquad {T}_I = f_{\text{Tokenizer}}(I),
\end{equation}
where ${T}_S \in \mathbb{N}^{L_S}$ and ${T}_I \in \mathbb{N}^{L_I}$ denote integer token sequences of length $L_S$ and $L_I$, respectively. 

\subsubsection{Reasoning and Planning}

At the core of our framework lies the reasoning and planning stage, which serves as the cognitive substrate for bridging high-level human intent and low-level embodied execution. While conventional approaches to humanoid planning often rely either on 
\emph{geometric search}, which emphasizes kinematic feasibility in configuration space, 
or on \emph{semantic understanding}, which interprets task goals at a symbolic level. Our framework advances beyond these paradigms by proposing {Embodied Chain-of-Action Reasoning}, as illustrated in Fig.~\ref{fig:COA},  which reconceptualizes humanoid planning as a cognitive process rather than a purely geometric or semantic one.  By leveraging foundation models, this approach explicitly grounds high-level instructions in perceived scene semantics and incrementally refines them into executable action chains. In doing so, it overcomes the brittleness of geometric planners and the abstraction gap of semantic interpreters, yielding plans that are both robust and transparent in their reasoning trace. Specifically, given the tokenized instruction ${T}_I$, the tokenized scene description ${T}_S$ derived from observations $O$, and the action library $L$, our reasoning model generates a structured sequence of actions by first producing intermediate reasoning states $R$ and then generating an executable action chain $A$. Formally, this process can be expressed as:
\begin{equation}
\resizebox{.8\columnwidth}{!}{$
\begin{aligned}
p(R, A \mid {T}_I, {T}_S, L) 
&= \prod_{i=1}^N p_\theta(R_i \mid {T}_I, {T}_S, L, R_{<i}) \\
&\quad \times \prod_{j=1}^{M} p_\theta(A_j \mid {T}_I, {T}_S, L, R, A_{<j}),
\end{aligned}
$}
\end{equation}
where $R = \{R_1, \dots, R_N\}$ denotes the sequence of intermediate reasoning states 
(e.g., affordance analysis, spatial inference, whole-body feasibility), 
and $A = \{A_1, \dots, A_M\}$ represents the sequence of grounded actions selected from the library $L$. 
The first part of the product computes the conditional probabilities of each reasoning step based on the accumulated prior reasoning, 
while the second part calculates the probabilities of action steps given both the prior reasoning and the previously generated actions.

\paragraph{Object Affordance Analysis}

In humanoid loco-manipulation, the feasibility of an action critically depends on the physical properties of objects. For example, grasping a lightweight cup differs fundamentally from attempting to lift a heavy box. Pure semantic reasoning is insufficient, as it cannot ensure that the generated actions are physically realizable. Affordance analysis addresses this challenge by explicitly modeling the interaction possibilities afforded by each object. Formally, we define the set of perceived objects as $\mathcal{E} = \{e_i\}_{i=1}^N$. Each object $e_i$ is associated with an affordance vector
\(
\mathbf{a}_i = [\alpha_i^{\text{size}}, \alpha_i^{\text{weight}}, \alpha_i^{\text{rigid}}, \alpha_i^{\text{mov}}, \alpha_i^{\text{act}}],
\)
where $\alpha_i^{(\cdot)} \in \mathbb{R}$ quantifies the corresponding property. The affordance vector $\mathbf{a}_i$ is integrated into the intermediate reasoning state as
\begin{equation}
R^{\text{aff}}_t = f_{\text{aff}}(e_i,\mathbf{a}_i,R_{<t}),
\end{equation}
ensuring that reasoning explicitly reflects physical feasibility. Action generation is then conditioned on the affordance-refined reasoning state:
\begin{equation}
p(A_j \mid T_I, T_S, L, R^{\text{aff}}_t) \propto 
\sum_{i=1}^{N} \phi(A_j, e_i, \mathbf{a}_i),
\end{equation}
where $\phi(\cdot)$ measures the compatibility between a candidate action $A_j$, an object $e_i$, and its affordance vector. In this way, object affordance analysis grounds reasoning in physical interaction constraints and prunes the hypothesis space to semantically consistent and physically feasible actions.  

\paragraph{Region Spatial Reasoning} 
Real-world environments are often cluttered and partially observable, where objects of interest may be occluded or outside the immediate field of view. If the agent reasons only over visible entities, it may fail to plan for critical targets hidden from direct perception. Spatial reasoning addresses this limitation by hypothesizing plausible locations for occluded or unseen objects based on contextual cues and semantic priors. To operationalize this, we define a set of candidate regions $\mathcal{R} = \{r_k\}_{k=1}^K$. Each region $r_k$ is assigned a weight $w_k$ derived from CLIP-based visual–semantic similarity between the cropped region and the target description, normalized via softmax. This produces a probabilistic spatial prior $\{(r_k, w_k)\}_{k=1}^K$. The reasoning state is then updated as
\begin{equation}
R^{\text{spatial}}_t = f_{\text{spatial}}\!\big(\mathcal{E}_{\text{obs}}(S), \{(r_k,w_k)\}, R_{<t}\big),
\end{equation}
where $f_{\text{spatial}}$ fuses observed entities, spatial priors, and reasoning history. The resulting bias on action generation is expressed as
\begin{equation}
p_\theta(A_j \mid T_I, T_S, L, R^{\text{spatial}}_t, A_{<j}) 
\propto \sum_{k=1}^{K} w_k\, \Phi_j(r_k,\cdot),
\end{equation}
where $\Phi_j(\cdot)$ measures the compatibility between a candidate action $A_j$ and hypothesized regions. Unlike affordance analysis, which imposes static feasibility constraints on individual actions, spatial reasoning requires sequential decision-making, and therefore depends on the history of previously generated actions $A_{<j}$. Through this mechanism, region spatial reasoning improves robustness in partially observable settings by directing exploration and action selection toward the most plausible hidden targets.

\paragraph{Body Movement Inference} 
Humanoid loco-manipulation requires not only reasoning about objects and spatial context but also ensuring that planned actions are consistent with the robot’s kinematic and dynamic embodiment. Without explicit modeling of bodily feasibility, an agent may propose actions that are semantically correct but physically infeasible, such as grasping an object outside its reachable workspace or lifting beyond its joint torque limits. Body movement inference addresses this challenge by systematically grounding action generation in the physical capabilities of the humanoid body. Formally, let $\mathcal{J} = \{j_m\}_{m=1}^M$ denote the set of controllable joints. Each joint $j_m$ is characterized by a feasibility descriptor
\(
\mathbf{b}_m = [\beta_m^{\text{range}}, \beta_m^{\text{torque}}, \beta_m^{\text{role}}],
\)
where $\beta_m^{\text{range}}$ encodes the kinematic range of motion, $\beta_m^{\text{torque}}$ specifies its dynamic limits, and $\beta_m^{\text{role}}$ denotes functional contributions (e.g., reaching, grasping, locomotion). 
The body-level reasoning state is updated as
\begin{equation}
R^{\text{body}}_t = f_{\text{body}}\!\big(\{(j_m,\mathbf{b}_m)\}_{m=1}^M, R_{<t}\big),
\end{equation}
which integrates joint feasibility with prior reasoning. 
Action generation is then constrained by
\begin{equation}
p_\theta(A_j \mid T_I, T_S, L, R^{\text{body}}_t, A_{<j}) 
\propto \sum_{m=1}^{M} \psi(A_j, j_m, \mathbf{b}_m),
\end{equation}
where $\psi(\cdot)$ measures the alignment between candidate actions and the kinematic/dynamic feasibility of the involved joints. As a result, body movement inference provides the body-level grounding of CoA reasoning, ensuring that action plans not only satisfy semantic and spatial constraints but also respect the physical embodiment of the humanoid, thereby guaranteeing safe and executable behavior.

\subsection{Execution and Control}

Following the reasoning and planning stage, the symbolic action chain $A$ must be grounded into concrete behaviors that can be executed by the humanoid robot. To achieve this, we define a structured action library $\mathcal{L}$ (Table~\ref{tab:humanoid_action_library}), which provides low-level primitives tailored for bipedal humanoid robots. These primitives are grouped into three categories: perception-guided actions for visually informed behaviors, locomotion actions for navigation and posture adjustment, and manipulation actions for physical interaction with objects. Together, they enable the agent to translate abstract plans into physically realizable movements.  Formally, let $A = \{A_1, \dots, A_M\}$ denote the sequence of symbolic actions produced during reasoning. Execution is modeled as a policy that maps these actions into low-level motor commands $\tau_{1:T}$:
\begin{equation}
p(\tau_{1:T} \mid A, \mathcal{L}) \;=\; \prod_{t=1}^{T} \pi_\theta(\tau_t \mid A, \mathcal{L}, \tau_{<t}),
\end{equation}
where $\tau_t$ denotes the joint torques or control signals at time step $t$, and $\pi_\theta$ represents the trained locomotion and manipulation controllers.  By grounding symbolic plans into embodied motor trajectories, Execution and Control closes the perception–reasoning–action loop, ensuring that the humanoid robot performs complex, long-horizon tasks in a safe and physically consistent manner.

\begin{table}[h!]
\centering
\caption{Action Library defining primitive skills for humanoid agent.}
\label{tab:humanoid_action_library}
\renewcommand{\arraystretch}{1.1} % tighter row spacing
\setlength{\tabcolsep}{4pt}       % tighter column spacing
\small
\resizebox{0.45\textwidth}{!}{
\begin{tabular}{ll}
\toprule
\textbf{Action} & \textbf{Description} \\
\midrule
% \multicolumn{2}{l}{\textbf{Perception}} \\
\texttt{FIND(object)}         & Detect and localize the target object \\ 
\midrule
% \multicolumn{2}{l}{\textbf{Locomotion}} \\
\texttt{MOVE(x,y)}            & Translate by $(x,y)$ in robot frame \\
\texttt{ROTATE(rZ)}           & Rotate counter-clockwise by $rZ$ \\
\texttt{STOP}                 & Halt locomotion and stabilize posture \\
\texttt{INCREASE\_HEIGHT}     & Raise standing height \\
\texttt{DECREASE\_HEIGHT}     & Lower standing height \\
\midrule
% \multicolumn{2}{l}{\textbf{Manipulation}} \\
\texttt{HOLD(object)}         & Maintain bi-manual grasp on object \\
\texttt{RELEASE(object)}      & Open gripper and disengage grasp \\
\texttt{GRASP(object, hand)}  & Close gripper on object (left/right/both) \\
\texttt{LIFT(object)}         & Raise object to chest height \\
\texttt{RAISE(object)}        & Bi-manual raise to chest level \\
\texttt{REARRANGE(object)}    & Move object to workspace region \\
\texttt{PUT(object, location)}& Release object at specified pose/location \\
\bottomrule
\end{tabular}}
\end{table}

\begin{table*}[b]
\centering
\caption{Comparison of manipulation performance across baselines and our method on simple and complex tasks. }
\resizebox{0.8\textwidth}{!}{
\begin{tabular}{l|cc|cc|cc|cc}
\toprule
\multirow{2}{*}{Task} & \multicolumn{2}{c|}{Loco-Manipulation Planning~\cite{murooka2021humanoid}} & \multicolumn{2}{c|}{Translated GPT3~\cite{huang2022language}} & \multicolumn{2}{c|}{LLM Behavior Planner~\cite{wang2024autonomous}}  & \multicolumn{2}{c}{Humanoid-COA (Ours)} \\
 & Executable & Success & Executable & Success & Executable & Success & Executable & Success \\ 
 % & Exec. &  Succ. & Exec. &  Succ. & Exec. &  Succ. & Exec. &  Succ. \\ 
 \midrule
% \multicolumn{9}{l}{\textbf{Simple}} \\ 
Object grasping & 86.6 & 80.0 & 80.0 & 70.0 & 90.0 & 83.3 & \textbf{100.0} & \textbf{96.6} \\
Object relocation & 86.6 & 76.6 & 73.3 & 60.0 & 86.6 & 80.0 & \textbf{96.6} & \textbf{93.3} \\
 \midrule
% \multicolumn{9}{l}{\textbf{Complex}} \\
% Stacking & - & - & - & - & - & - & 85\% & 75\% \\
Spatial placement & 73.3 & 56.6 & 60.0 & 33.3 & 80.0 & 63.3 & \textbf{83.3} & \textbf{76.6} \\
Sequential manipulation & 83.3 & 73.3 & 63.3 & 53.3 & 83.3 & 73.3 & \textbf{93.3} & \textbf{86.6} \\
Rearrangement & 56.6 & 36.6 & 30.0 & 0.0 & 63.3 & 40.0 & \textbf{80.0} & \textbf{73.3} \\ 
\bottomrule
\end{tabular}
}
\label{tab:manipulation}
% \vspace{-7pt}
\end{table*}

%% file: sec/4_experiments.tex
\section{EXPERIMENTS}

\begin{figure*}[h]
\centering
\includegraphics[width=0.8\textwidth]{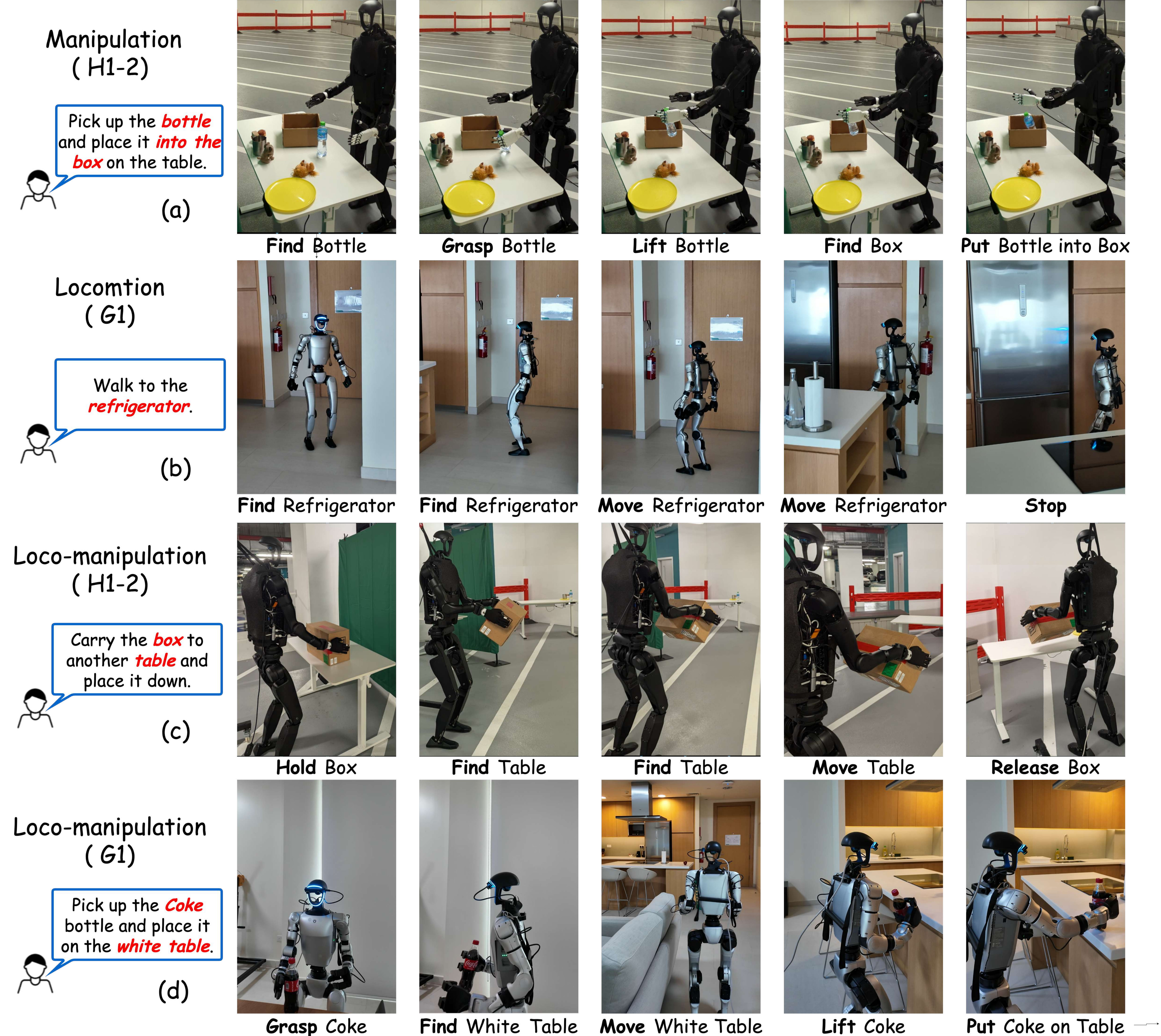}
% \captionof{figure}{Real-world qualitative results on three humanoid loco-manipulation tasks.}
\captionof{figure}{Real-world humanoid loco-manipulation tasks performed by two robots, Unitree H1-2 and G1, in two different scenarios: an open area and an apartment environment. Each task is specified by a human instruction (left), and the robot executes the corresponding action sequence to complete it (right), covering manipulation, locomotion, and integrated loco-manipulation.} \vspace{-20pt}
\label{fig:example}
\end{figure*}

\subsection{Experimental Setup}

\subsubsection{Robotic Platforms and Environments} We conducted real-world experiments to evaluate the proposed humanoid agent framework, using two humanoid robots, the Unitree H1-2 and the Unitree G1,  in two distinct environments: (i) an open-field setting populated with randomly placed objects and tables to emulate unstructured scenarios, and (ii) a standardized apartment environment consisting of one bedroom, one bathroom, one kitchen, and one living room. The H1-2 is a full-sized bipedal humanoid robot featuring 27 degrees of freedom and Inspire Hand 6-DoF grippers for dexterous manipulation.  The G1 is a smaller humanoid platform designed with 21 degrees of freedom and a pair of adaptive grippers to support everyday manipulation tasks. Both robots are equipped with two RGB-D cameras: a head-mounted, downward-facing camera for fine-grained manipulation, and a chest-mounted, forward-facing camera for navigation and object search. For both robots, manipulation is executed through end-effector control with inverse kinematics used to compute joint configurations, while locomotion is managed through high-level navigation commands. Communication across all system components is coordinated via the Data Distribution Service middleware.

\subsubsection{Implementation Details} We adopt GPT-4V~\cite{2023gpt4} as the VLM to process RGB-D observations and generate scene descriptions, and employ GPT-4 as the LLM to construct reasoning chains for long-horizon task planning. For loco-manipulation, we further integrate a trained policy~\cite{h2-compact} that enables the robot to maintain stability under external loads during walking and manipulation. After each action execution, the system updates the action list by integrating new observations with previously generated reasoning chains and action sequences. This iterative feedback loop enables real-time plan adaptation and improves robustness under dynamic environmental changes.

\subsubsection{Baslines and Evaluation Metrics} We compare our framework against three representative baselines: 
(i) \textbf{Loco-Manipulation Planning}~\cite{murooka2021humanoid}, which employs graph search with reachability maps; 
(ii) \textbf{Translated GPT3}~\cite{huang2022language}, which extracts actionable knowledge from language models for zero-shot planning; and 
(iii) \textbf{LLM Behavior Planner}~\cite{wang2024autonomous}, which leverages grounded language models for humanoid loco-manipulation.  
Following~\cite{wang2024autonomous}, performance is assessed using two complementary metrics: 
\textbf{Executable}, which verifies whether the generated action sequences are syntactically valid and executable on the humanoid robot, ensuring consistency with the action library; and 
\textbf{Success}, which measures whether the executed sequences accomplish the intended task goals, thereby reflecting task-level accuracy and reliability.

\subsubsection{Task Design} To systematically evaluate the proposed framework, we designed three categories of tasks: manipulation, locomotion, and loco-manipulation. Each category was instantiated at two levels of difficulty. The simple tasks involve single-step objectives, such as grasping an object or moving towards a target location. In contrast, the complex tasks require multi-step reasoning and action sequencing, such as combining object manipulation with spatial relocation or sequential navigation across multiple landmarks. This design allows us to assess not only the framework’s basic capability in perception and control but also its robustness and scalability in long-horizon, unstructured scenarios.

% \begin{figure}[h]
% \centering
% \includegraphics[width=0.5\textwidth]{imgs/results.drawio.png}
% % \captionof{figure}{Real-world qualitative results on three humanoid loco-manipulation tasks.}
% \captionof{figure}{Real-world qualitative results on three humanoid loco-manipulation tasks. In the \textbf{top row (Loco-Manipulation)}, the robot holds a box, locates the target, walks to it, and places the box. The \textbf{middle row (Object Manipulation)} the robot identifies a bottle, grasps it, lifts it, and places it. In the \textbf{ bottom row (Rearrangement and Manipulation)} robot locates multiple objects, rearranges them, and places the selected object in the designated position.}
% \label{fig:humanoid_tasks}
% \end{figure}

\subsection{Experimental Results}
\subsubsection{Manipulation} To facilitate systematic evaluation, we classify manipulation tasks into simple and complex categories. {Simple tasks} include \textit{Object Grasping} (picking up a single object) and \textit{Object Relocation} (moving it to a nearby position without strict constraints). {Complex tasks} include \textit{Spatial Placement} (positioning one object relative to another), \textit{Sequential Manipulation} (performing consecutive operations on multiple objects), and \textit{Rearrangement} (structuring several objects into a desired configuration). Results are summarized in Table~\ref{tab:manipulation}. On simple tasks, our method achieves near-perfect executability and success, substantially outperforming prior approaches. The advantage becomes more evident in complex tasks requiring multi-step reasoning and coordination. For example, in \textit{Spatial Placement} and \textit{Sequential Manipulation}, our framework yields much higher success rates than LLM Behavior Planner~\cite{wang2024autonomous}. Even on the most challenging \textit{Rearrangement} task, it consistently surpasses all baselines, highlighting the robustness of CoA reasoning in long-horizon scenarios. An example is shown in Fig.~\ref{fig:example}(a), where the H1-2 robot follows a natural language instruction to pick up a bottle and place it into a box. Overall, these results validate that affordance grounding, spatial inference, and whole-body feasibility jointly ensure reliable execution in manipulation tasks.

\begin{table*}[h]
\centering
\caption{Comparison of Locomotion performance across baselines and our method on simple and complex tasks. }
\resizebox{0.8\textwidth}{!}{
\begin{tabular}{l|cc|cc|cc|cc}
\toprule
\multirow{2}{*}{Task} & \multicolumn{2}{c|}{Loco-Manipulation Planning~\cite{murooka2021humanoid}} & \multicolumn{2}{c|}{Translated GPT3~\cite{huang2022language}} & \multicolumn{2}{c|}{LLM Behavior Planner~\cite{wang2024autonomous}}  & \multicolumn{2}{c}{Humanoid-COA (Ours)} \\
 & Executable & Success & Executable & Success & Executable & Success & Executable & Success \\ 
 % & Exec. &  Succ. & Exec. &  Succ. & Exec. &  Succ. & Exec. &  Succ. \\ 
 \midrule
% \multicolumn{9}{l}{\textbf{Simple}} \\ 
Target approach & 90.0 & 80.0 & 86.6 & 66.6 & 93.3 & 83.3 & \textbf{100.0} & \textbf{96.6} \\
Pose adjustment & 83.3 & 70.0 & 76.6 & 60.0 & 83.3 & 66.6 & \textbf{100.0} & \textbf{90.0} \\
 \midrule
% \multicolumn{9}{l}{\textbf{Complex}} \\
Sequential navigation & 70.0 & 60.0 & 63.3 & 43.3 & 73.3 & 56.6 & \textbf{93.3} & \textbf{86.6} \\
Navigation under occlusion & 46.6 & 16.6 & 0.0 & 0.0 & 46.6 & 26.6 & \textbf{80.0} & \textbf{63.3} \\
Long-horizon relocation & 36.6 & 13.0 & 0.0 & 0.0 & 33.3 & 13.3 & \textbf{66.6} & \textbf{56.6} \\
\bottomrule
\end{tabular}
} \vspace{-10pt}
\label{tab:locomotion}

\end{table*}

\begin{table*}[h]
\centering
\caption{Comparison of Loco-manipulation performance across baselines and our method on simple and complex tasks. }
\resizebox{0.8\textwidth}{!}{
\begin{tabular}{l|cc|cc|cc|cc}
\toprule
\multirow{2}{*}{Task} & \multicolumn{2}{c|}{Loco-Manipulation Planning~\cite{murooka2021humanoid}} & \multicolumn{2}{c|}{Translated GPT3~\cite{huang2022language}} & \multicolumn{2}{c|}{LLM Behavior Planner~\cite{wang2024autonomous}}  & \multicolumn{2}{c}{Humanoid-COA (Ours)} \\
 & Executable & Success & Executable & Success & Executable & Success & Executable & Success \\ 
 % & Exec. &  Succ. & Exec. &  Succ. & Exec. &  Succ. & Exec. &  Succ. \\ 
 \midrule 
Mobile Pick & 90.0 & 73.3 & 63.3 & 43.3 & 96.6 & 83.3 & \textbf{96.6} & \textbf{90.0} \\
Mobile Place & 93.3 & 83.3 & 66.6 & 50.0 & 96.6 & 86.6 & \textbf{100.0} & \textbf{96.6} \\
 \midrule
Sequential Loco-Manipulation  & 80.0 & 66.6 & 50.0 & 26.6 & 76.6 & 63.3 & \textbf{90.0} & \textbf{83.3} \\
Occlusion-Aware Loco-Manipulation & 60.0 & 43.3 & 0.0 & 0.0 & 63.3 & 46.6 & \textbf{80.0} & \textbf{66.6} \\
Long-Horizon Loco-Manipulation & 40.0 & 26.6 & 0.0 & 0.0 & 40.0 & 23.3 & \textbf{70.0} & \textbf{63.3} \\
\bottomrule
\end{tabular}
} \vspace{-20pt}
\label{tab:loco-manipulation}
\end{table*}

\subsubsection{Locomotion} In addition to manipulation, we also consider locomotion tasks, which are categorized into simple and complex cases. {Simple tasks} include \textit{Target Approach}, which requires navigating towards a specified object or landmark, and \textit{Pose Adjustment}, which involves refining the robot’s stance or orientation once in proximity to the target. {Complex tasks} comprise \textit{Sequential Navigation}, where the robot must visit multiple waypoints in a prescribed order, \textit{Navigation under Occlusion}, which requires reasoning about partially visible or hidden targets and exploring the environment to locate them, and \textit{Long-horizon Relocation}, involving extended traversal across distant regions to reach the goal position. Table~\ref{tab:locomotion} reports the results. For the simple tasks, our framework achieves perfect executability and above 90\% success, clearly outperforming prior methods. In the complex settings, the gap with baselines becomes more pronounced: for \textit{Sequential Navigation}, our method sustains robust performance (86.6\% success) while others degrade sharply; for \textit{Navigation under Occlusion}, embodied reasoning enables reliable target recovery under partial observability, leading to more than double the success rate of competing approaches. Even in the long-horizon case, our framework preserves significantly higher robustness than existing planners. As illustrated in Fig.~\ref{fig:example}(b), the G1 robot successfully follows a human instruction to walk to a refrigerator and stop in front of it, demonstrating the effectiveness of our approach on real-world locomotion tasks. These findings demonstrate that explicit spatial reasoning and iterative feedback are crucial for long-horizon and occlusion-aware humanoid locomotion.

\subsubsection{Loco-Manipulation} Beyond isolated manipulation and locomotion, we further evaluate our framework on integrated loco-manipulation tasks, which couple mobility with object interaction. \textit{Simple tasks} include \textit{Mobile Pick}, where the robot navigates to and grasps a specified object, and \textit{Mobile Place}, which requires carrying an object to a designated location. \textit{Complex tasks} comprise \textit{Sequential Loco-Manipulation}, involving multi-step navigation and manipulation in sequence, \textit{Occlusion-Aware Loco-Manipulation}, which requires locating and manipulating objects under partial observability, and \textit{Long-Horizon Loco-Manipulation}, combining extended navigation with precise manipulation at distant targets.  The results in Table~\ref{tab:loco-manipulation} show that our framework consistently achieves the highest executability and success rates across both simple and complex cases. On simple tasks, our method exceeds 90\% success, outperforming all baselines. For complex scenarios, the benefits of embodied CoA are more evident: in \textit{Sequential Loco-Manipulation}, our approach achieves 83.3\% success compared to 63.3\% for the strongest baseline, while in \textit{Occlusion-Aware Loco-Manipulation}, our method maintains 66.6\% success, substantially higher than others which largely fail under occlusion. Even in the most demanding \textit{Long-Horizon Loco-Manipulation}, our framework preserves over 60\% success, more than doubling the performance of baseline planners. Figures~\ref{fig:example}(c) and (d) show successful executions of loco-manipulation tasks, where the robots follow human instructions to carry a box to another table and place it down, or to pick up a Coke bottle and place it on a white table, demonstrating the effectiveness of our framework in real-world scenarios. These results highlight the strength of our reasoning in coordinating perception, locomotion, and manipulation for long-horizon, unstructured humanoid tasks.

\subsection{Discussion}

\subsubsection{Action Library Evaluation} Table~\ref{tab:action_library} reports the performance of primitive actions in the humanoid action library. Most skills, such as \textit{FIND}, \textit{MOVE}, \textit{GRASP}, and \textit{PUT}, achieve near-perfect executability and success, confirming their robustness. Simpler actions like \textit{RELEASE} are executed quickly, whereas complex ones such as \textit{MOVE} and \textit{REARRANGE} take longer and show reduced reliability, with \textit{REARRANGE} being the most error-prone. These results verify the soundness of the action set but also reveal its limits, motivating the need for our proposed {embodied CoA reasoning} to dynamically compose primitives for robust loco-manipulation in unstructured environments.

\begin{table}[h]
\centering
\caption{Performance of individual actions in the action library.}
\label{tab:action_library}
\begin{tabular}{lccc}
\toprule
Task       & Executable (\%) & Success (\%) & Time (s) \\ 
\midrule
FIND       & 100.0 & 97.1 & 11.2 \\
MOVE       & 100.0 & 96.6 & 24.6 \\
HOLD       &  95.0 & 88.0 & 12.9\\
RELEASE    &  97.1 & 96.6 & 1.3 \\
GRASP      & 100.0 & 96.6 & 11.7 \\
LIFT       & 100.0 & 100.0 & 8.4 \\
RAISE      & 100.0 & 100.0 & 9.1 \\
REARRANGE  &  80.0 & 73.3 & 23.0 \\
PUT        & 100.0 & 94.9 & 11.7 \\
\bottomrule
\end{tabular}
\end{table}

\subsubsection{Ablation Study of Chain-of-Action Reasoning} To evaluate the contribution of each component in our proposed CoA reasoning framework, we conducted an ablation study by progressively enabling \textit{Object Affordance Analysis}, \textit{Region Spatial Reasoning}, and \textit{Whole-Body Movement Inference}.  As shown in Table~\ref{tab:ablation}, removing all three modules leads to poor performance, with only 50\% executability and 45\% success rate, indicating that action sequences generated without structured reasoning are often ill-formed or fail to complete the tasks.  Introducing \textit{Object Affordance Analysis} substantially improves executability to 75\% and success to 65\%,  demonstrating its role in constraining actions to physically feasible interactions.  Adding \textit{Region Spatial Reasoning} further boosts performance to 85\% executability and 70\% success,  showing its effectiveness in guiding exploration under occlusion.  Finally, incorporating all three modules, including \textit{Whole-Body Movement Inference}, achieves the best performance (90\% executability and 75\% success), highlighting the importance of reasoning over kinematic feasibility for reliable execution.  These results confirm that each stage of the CoA reasoning pipeline contributes complementary benefits, and their integration yields the most robust and accurate action planning.

\begin{table}[h]
\centering
\caption{AAblation study of the proposed Chain-of-Action Reasoning mechanism on executability and success rate.}
\resizebox{0.5\textwidth}{!}{
\begin{tabular}{>{\centering\arraybackslash}p{3cm}|>{\centering\arraybackslash}p{2.5cm}|>{\centering\arraybackslash}p{3cm}|cc}
\toprule
Object Affordance Analysis & Region Spatial  Reasoning  & Whole-Body Movement Inference & Executable  & Success  \\ 
\midrule
\ding{55} &  \ding{55} &   \ding{55}  &   50\%  &  45\%              \\
\ding{51} &  \ding{55} &   \ding{55}  &   75\%  &  65\%             \\
\ding{51} &  \ding{51} &   \ding{55}  &   85\%  &  70\%             \\
\ding{51} &  \ding{51} &   \ding{51}  &   90\%  &  75\%           \\ 
\bottomrule
\end{tabular}} \vspace{-20pt}
\label{tab:ablation}
\end{table}

%% file: sec/5_conclusion.tex
\section{CONCLUSION}
In this work, we addressed the longstanding challenge of humanoid loco-manipulation, which requires coordinating whole-body locomotion with dexterous manipulation under natural language instructions. We introduced the first humanoid agent framework, Humanoid-COA, which integrates foundation model reasoning with an embodied CoA mechanism, enabling the decomposition of high-level human intent into executable whole-body behaviors. Our design follows the perception–reasoning–action paradigm, where the reasoning stage is realized through a CoA process that combines affordance grounding, spatial inference, and whole-body feasibility analysis to ensure robust execution in unstructured environments. Extensive experiments on two humanoid robots, Unitree H1-2 and G1, demonstrated the effectiveness of our approach across manipulation, locomotion, and loco-manipulation tasks. The framework consistently achieved higher executability and success rates than prior methods, with especially pronounced improvements in long-horizon and occlusion-aware scenarios.